\documentclass[letterpaper, 10 pt, conference]{ieeeconf}  

\IEEEoverridecommandlockouts                              

\usepackage{graphicx}
\usepackage{multirow}
\usepackage{amssymb}
\usepackage{amsmath}
\usepackage{amstext}
\usepackage{comment}
\usepackage[dvipsnames]{xcolor}
\usepackage{pifont}
\usepackage{array}
\newcommand{\PreserveBackslash}[1]{\let\temp=\\#1\let\\=\temp}
\newcolumntype{C}[1]{>{\PreserveBackslash\centering}p{#1}}
\newcolumntype{R}[1]{>{\PreserveBackslash\raggedleft}p{#1}}
\newcolumntype{L}[1]{>{\PreserveBackslash\raggedright}p{#1}}
\usepackage[colorlinks=true,citecolor=blue,urlcolor=blue]{hyperref}
\usepackage{cite}
\usepackage{overpic}
\usepackage{textcomp}

\title{\LARGE \bf
Efficient Physically-based Simulation of Soft Bodies in \\ Embodied Environment for Surgical Robot}

\author{
Zhenya Yang, Yonghao Long, Kai Chen, Wang Wei, Qi Dou \\
\thanks{Z. Yang, Y. Long, K. Chen, W. Wei and Q. Dou are with the Department of Computer Science and Engineering, The Chinese University of Hong~Kong.}
\thanks{\textit{Corresponding author: Qi Dou (qidou@cuhk.edu.hk).}}%
}
\begin{document}
\maketitle
\thispagestyle{empty}
\pagestyle{empty}
\begin{abstract}

Surgical robot simulation platform plays a crucial role in enhancing training efficiency and advancing research on robot learning.
Much effort have been made by scholars on developing open-sourced surgical robot simulators to facilitate research. 
We also developed SurRoL formerly, an open-source, da Vinci Research Kit (dVRK) compatible and interactive embodied environment for robot learning. 
Despite its advancements, the simulation of soft bodies still remained a major challenge within the open-source platforms available for surgical robotics. 
To this end, we develop an interactive physically based soft body simulation framework and integrate it to SurRoL. 
Specifically, we utilized a high-performance adaptation of the Material Point Method (MPM) along with the Neo-Hookean model to represent the deformable tissue. 
Lagrangian particles are used to track the motion and deformation of the soft body throughout the simulation and Eulerian grids are leveraged to discretize space and facilitate the calculation of forces, velocities, and other physical quantities. 
We also employed an efficient collision detection and handling strategy to simulate the interaction between soft body and rigid tool of the surgical robot. 
By employing the Taichi programming language, our implementation harnesses parallel computing to boost simulation speed. 
Experimental results show that our platform is able to simulate soft bodies efficiently with strong physical interpretability and plausible visual effects.
These new features in SurRoL enable the efficient simulation of surgical tasks involving soft tissue manipulation and pave the path for further investigation of surgical robot learning.
The code will be released in a new branch of SurRoL github repo.

\end{abstract}

\section{Introduction}

Robotic surgery, owing to its significant advantages of flexibility and dexterity, has been increasingly and widely adopted.
How to automate surgical tasks is a hot topic in recent research.
Embodied AI is a powerful method that enables surgical robots to learn policies from perceiving, understanding, and acting in the environment, which empowers robots to take action and make decision autonomously.
In most cases, policies are trained in simulator rather than the real world because Embodied AI algorithms typically require millions of interactions with the environment, which can be time-consuming, expensive, and risky in the real world\cite{scheikl2022sim}.

\begin{figure}[t]
\centering
   \begin{overpic}[width=\columnwidth]{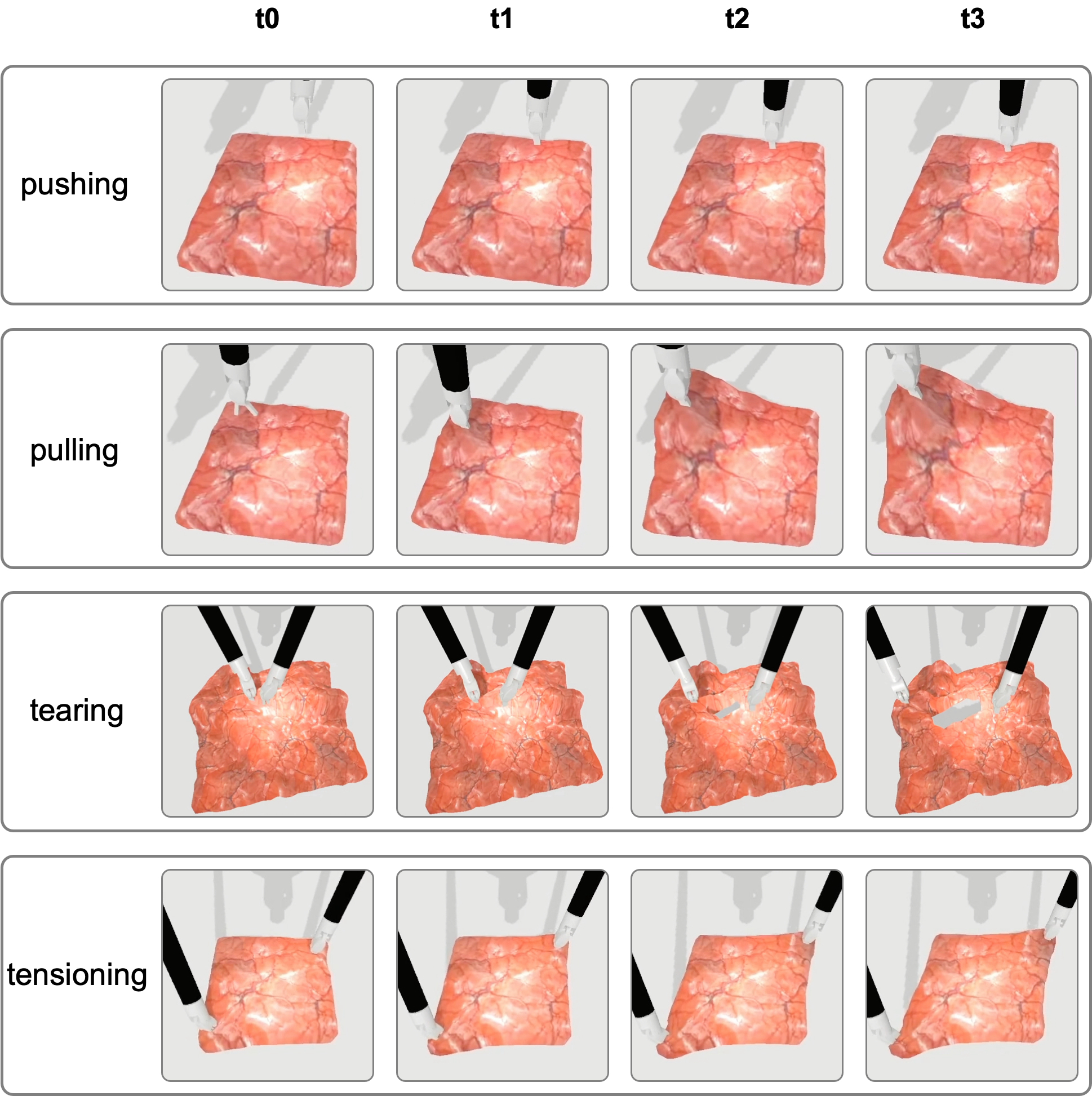}
    \end{overpic}
    \caption{\textbf{The motion sequences of four typical actions for soft body manipulation}: pushing\cite{marban2019recurrent}, pulling\cite{scheikl2022sim}, tearing\cite{wolper2020anisompm} and
    tensioning\cite{ou2023sim}. 
    All results are simulated and rendered from SurRoL~\cite{xu2021surrol,long2022integrating,long2023humanintheloop,huang2023demonstrationguided,huang2023valueinformed} using 30K particles.}
    \label{fig::teaser}
    \vspace{-1em}
\end{figure}

\begin{figure*}[t]
   \centering
   \vspace{8pt}
   \begin{overpic}[width=\textwidth]{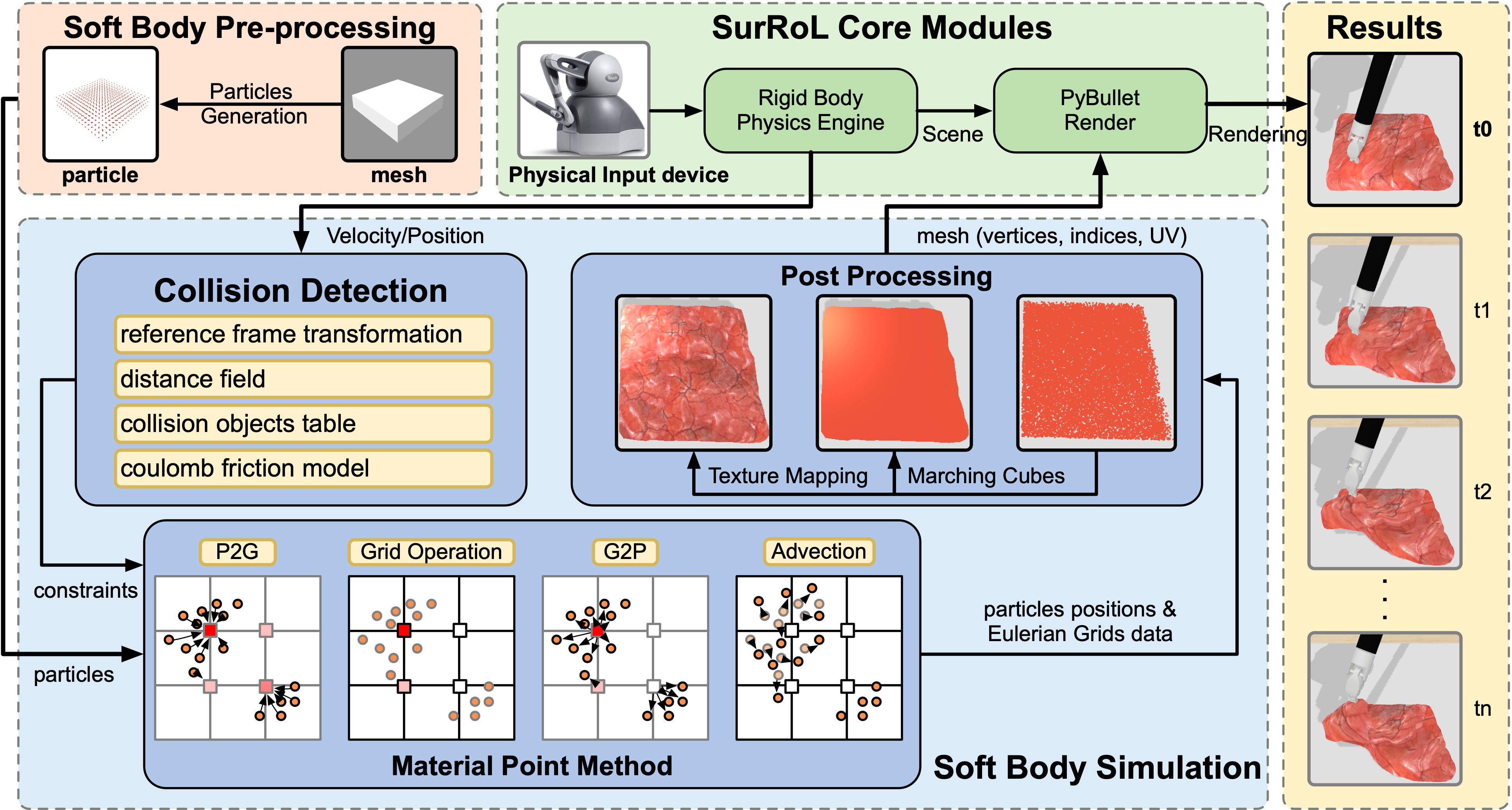}
    \end{overpic}
    \caption{\textbf{Overview of the System Design.} Material point method based soft body simulator is implemented as an extension (blue) of original SurRoL (green)\cite{xu2021surrol}.
    Lagrangian particles are generated from mesh or distribution functions before the simulation (red).
    At each simulation step, the soft body simulator receives the kinematic information of the surgical robot from the physics engine of original SurRoL and performs collision detection. 
    The MPM algorithm receives the collision constraints and performs the simulation, and the results are processed to enhance visual effects.
    The processed geometry and visual data are then sent back to the PyBullet render to obtain the rendering results (yellow).
    }
    \vspace{-1em}
    \label{fig::sys_design}
\end{figure*}

A good simulated environment is key to policy learning.
However, providing physically based soft body simulation in surgical robot simulation platform is challenging.
On the one hand, physically based simulation algorithms involve knowledge about continuum mechanics\cite{gurtin1982introduction} which means it is more complex than rigid body simulation.
On the other hand, how to efficiently integrate soft body simulation functionality into robot simulation platform is an engineering challenge.
dVRL\cite{dVRL}, the first open-source reinforcement learning (RL) environment developed for surgical robot, only supports rigid body simulation.
Xu et al. proposed SurRoL\cite{xu2021surrol,long2022integrating,long2023humanintheloop,huang2023demonstrationguided,huang2023valueinformed}, a dVRK\cite{dvrk} compatible platform for surgical robot learning which shares the same limitation. 
Munawar et al. implemented a simulator for surgical robot learning called Asynchronous Multi-Body Framework (AMBF)\cite{munawar2019asynchronous,ambf,suturing_ambf}.
The soft body simulation of AMBF relies on the Bullet Physics library\cite{PyBullet} for simulation using position based dynamics (PBD)\cite{MULLER_pbd}. 
Tagliabue et al. proposed UnityFlexML\cite{tagliabue-sim}, which was built on Unity and utilized NVIDIA FleX\cite{macklin2014unified} as its physics engine.
NVIDIA FleX is a particle based library also using PBD algorithm to simulate the deformable objects. 
PBD is fast and easy to implement, but it was not derived from physical rules which makes it lack physical interpretability.

We extended the robot learning platform SurRoL\cite{xu2021surrol} by integrating a physically based and efficient soft body simulation algorithm.
In this work, we implemented Material Point Method (MPM)\cite{SULSKY_mpm}, a hybrid Lagrangian-Eulerian method which can be used to simulate deformable objects.
The simulation algorithm was implemented using Taichi programming language\cite{taichi}, a domain-specific language embedded in Python that helps users easily write portable, high-performance parallel programs.
Taichi helps make the most use of hardware and boost simulation performance greatly in a user-friendly way.
A collision detection module was introduced to connect original rigid physics engine and newly developed soft body simulator, helping simulate the contact and interaction between surgical tools and biological tissue.
We employed an optimized collision detection method to avoid redundant Signed Distance Field (SDF) computations which helps simulation achieve interactive speeds.
Post processing of simulation results is utilized in our simulator to enhance the visual effects.
The overall system design is illustrated in Fig. \ref{fig::sys_design}.
Our work provides an interactive embodied environment which supports physically based soft body simulation for surgical robot learning.
This simulator can be used to generate large amounts of training data of soft tissue manipulation for policy learning which is important for surgical robot tasks automation.
Our embodied environment will be open source and we hope that our platform will serve as a convenient tool for researchers interested in surgical robot learning and medical simulation.

\section{Related Works}

\subsection{Soft Body Simulation}
Soft body simulation is more complex than rigid body 
simulation due to the deformation of the material. 
There are various methods that could be used to simulate soft bodies for different goals. 
One commonly used simulation algorithm in game is 
Position Based Dynamics (PBD)\cite{MULLER_pbd}. 
PBD operates directly on positions, estimating vertices' positions and manipulating them by solving constraints. 
Then, it updates velocity of each vertex by using positions and starts next simulation step. 
PBD is known for its efficiency and plausible visual results\cite{macklin2016xpbd}, but it lacks strong physical interpretability\cite{yin2021modeling}. Therefore, it is more suitable for applications where high physical accuracy is not required. 
Another simulation algorithm is Finite Element Method (FEM)\cite{in2fem}. 
FEM divides a continuous domain into a set of sub-domains, combines all element equations into a global equations system and solves it finally. 
FEM has high physical fidelity and interpretability. However, the computation of FEM is expensive, making it challenging to apply in large-scale real-time applications.

The Particle-in-cell (PIC) method was first developed by computational physicists for plasma simulation.
The original PIC method is stable but dissipative.
FLIP\cite{FLIP}, an extension of PIC method used to simulate the fluid, was designed to solve the dissipation but it is unstable and noisy. 
Material Point Method (MPM)\cite{SULSKY_mpm} is an extension of FLIP and has been used in the production of visual effects in movies, such as the snow\cite{stomakhin-snow-mpm} in \textit{Frozen}, because of its simulation realism.
MPM is a hybrid Lagrangian-Eulerian method which combines the Lagrangian particles and Eulerian grids together. 
Compared to FEM, MPM is more suitable for modelling large material deformations and handling self-collision because of the use of Eulerian grids.
Moving Least Square Material Point Method (MLS-MPM)\cite{mls-mpm} is an improved variant of MPM which utilizes Affine Particle-In-Cell Method (APIC)\cite{jiang2015affine} as transfer and avoids the costly B-spline kernel gradients computation. 
MLS-MPM achieves a good balance between simulation realism and efficiency.
Owing to these advantages, we finally opted for MLS-MPM as the core algorithm of our soft body simulator.

\subsection{Surgical Robot Simulation Platform}
Some software products such as SIMBIONIX\cite{bar2006simbionix} and VIRTAMED are designed for doctors’ training and education. 
However, these commercial software are closed source which makes it hard to access the simulation methods they used.
There are also several open-source surgical simulators designed for research.
Fontanelli et al. \cite{vrep} developed a dVRK\cite{dvrk} simulator based on V-REP and
Fan et al. \cite{fan2022unity} developed a dVRK simulator based on Unity.
However, both of them are focused on surgical robot simulation and do not support learning algorithms.
da Vinci Reinforcement Learning (dVRL)\cite{dVRL} is the first open-source RL environment designed for research on autonomous surgery.
SurRoL\cite{xu2021surrol} is an open-source RL centered and dVRK compatible platform for surgical robot learning.
It uses PyBullet\cite{PyBullet} as its physics engine, and supports ten unique training tasks.
Surgical Gym\cite{surgicalgym} leveraged Isaac Gym\cite{isaac} to simulate surgical robot for efficient RL training.
However, all above simulators only support rigid-body simulation currently.

Tagliabue et al. introduced UnityFlexML\cite{tagliabue-sim}, a Unity based simulator to simulate the surgical task of soft tissue retraction.
It relies on NVIDIA FleX\cite{macklin2014unified}, a PBD-based simulation framework, to simulate the soft body. 
However, such method lacks physical interpretability and accuracy.
Munawar et al. developed an open-source simulation environment for surgical robot named Asynchronous Multi-Body Framework (AMBF)\cite{munawar2019asynchronous,ambf,suturing_ambf}, which supports deformable objects simulation.
The soft body simulation in AMBF relies on Bullet\cite{PyBullet} using position based dynamics (PBD)\cite{MULLER_pbd}.
However, due to the limited support for soft body simulation in Bullet (for example, the complexity of tuning and the numerical instabilities of deformable object simulation\cite{scheikl2023lapgym}), it can only support very limited deformable objects interaction.
LapGym\cite{scheikl2023lapgym} utilized the Simulation Open Framework Architecture (SOFA)\cite{sofa} as its physics engine for finite element method (FEM) simulation of deformable objects.
While it achieved promising results, the simulation of complex deformations with large volume in LapGym is computationally demanding. 
In this paper, we extended the SurRoL platform with physically based soft body simulation based on MPM, enabling our surgical robot learning platform to efficiently simulate the soft body.

\section{Methods}
In this section, we begin by introducing the design of our newly extended embodied environment, highlighting the intercommunication between the newly developed modules for soft body simulation and the original simulation platform.
The soft body simulation algorithm we used to simulate biological tissue will be illustrated in \ref{mpm_method} in detail.
To couple the newly implemented soft body simulation module and original rigid body simulation engine of PyBullet, we proposed an collision detection and handling strategy (in \ref{collision}) which helps connect these two parts efficiently.

\subsection{System Design}
To make the least change of the original framework and integrate the soft body simulation module into original platform seamlessly, the soft simulation was implemented as a separate module which only takes necessary information from original SurRoL as input and returns simulation results such as points positions or triangular meshes data to PyBullet for visualization. 
The whole system design is shown in Fig.~\ref{fig::sys_design}.
At first, the particles are generated from triangular mesh objects or distribution functions and sent to the soft body simulator.
The touch haptic device has been integrated into SurRoL to control the motion of surgical robot.
At each simulation step, the soft body simulator takes the kinematic information of surgical robot from original SurRoL to update the Distance Field and Collision Objects Table which will be used to detect collision between grid nodes and surgical robot.
The details of collision detection are illustrated in Fig. \ref{fig::collision}.
Then, the MPM algorithm is performed to simulate the soft tissue, and collisions are handled at the Grid Operation stage.
The rigid simulation of surgical robot is performed at the same time in Bullet physics engine\cite{PyBullet}.
After updating the particles positions, the post processing procedure is executed to enhance the visual effect.
To be more precise, the particles are first transferred to mesh surface by utilizing Marching Cubes algorithm\cite{marching_cubes}.
Then, the UV coordinates are computed for each vertex according to their positions.
These generated vertices, indices and UV coordinates are finally passed into PyBullet to generate the final rendering results of the simulation.
 

\subsection{Material Point Method}\label{mpm_method}
The physical interpretability and efficiency are two crucial criteria for soft body simulation in a robotics learning platform.
The main reason for choosing Material Point Method is its combination of physical interpretability and capability to model the large tissue deformations that commonly occur during surgeries.
However, the classical MPM is still not fast enough and expensive in storage.
Fortunately, advancements have been made in both the algorithm itself and hardware acceleration techniques for MPM.
Moving Least Square Material Point Method (MLS-MPM)\cite{mls-mpm} is a successful variant of MPM which not only avoids the costly B-spline kernel gradients but also make a good use of principle of locality in the program, leading to a great increase in performance.
In this paper, we implemented MLS-MPM to simulate soft tissue because of its high performance.
The Neo-Hookean model is used as constitutive model to predict elastic deformation of soft tissue because of its simplicity and computation efficiency.
Furthermore, it is able to capture the behavior of biological tissue effectively.
The entire simulation algorithm was written in Python and Taichi programming language\cite{taichi} which helps make the most use of hardware to boost efficiency.

The MPM algorithm can be divided into four stage: particle to grid (P2G), grid operation, grid to particle (G2P) and advection, as shown in Fig. \ref{fig::sys_design}. 
In MPM, Lagrangian particles represent material points that move with the biological tissue. 
These particles carry properties such as mass, velocity, and deformation. 
They are used to track the motion and deformation of the tissue throughout the simulation. 
Lagrangian particles provide a Lagrangian perspective, meaning they move with the material, allowing for accurate representation of tissue deformation and interaction with surgical tools.
Eulerian grids, on the other hand, are fixed grids that cover the entire simulation domain. 
These grids are used to discretize space and provide a framework for solving the governing equations of motion and other physical quantities. 
Eulerian grids are stationary and do not move with the material.
They facilitate the calculation of forces, velocities, and other quantities at each grid point.
The mathematics notations that will be used in this section are listed in Table \ref{tab:mpm_notations}.
The detailed derivation could be found at\cite{mls-mpm} and \cite{mpm_course}.
\begin{table}[]
    \centering
    \vspace*{5pt}
    \caption{\textbf{Notations List.}}
    \begin{tabular}{c|c|l}
    \hline
    \textbf{Variable}&\textbf{Type}  &\textbf{Meaning}\\
    \hline
    $\mathbf{v_i}$ & vector  & velocity of node i\\
    $\mathbf{v_p}$ & vector  & velocity of particle p\\
    $\mathbf{x_p}$ & vector  & position of particle p\\
    $\mathbf{C_p}$ & matrix  & velocity gradient of p\\
    $w_{ip}$ & scalar  & weight of node i and particle p\\
    $\triangle t$ & scalar  & timestep\\
    $\mathbf{f_{ip}}$ & vector  & stress between node i and particle p\\
    $\triangle x$ & scalar  & interval between each grid node\\
    $\mathbf{F_p}$ & matrix  & deformation gradient of particle p\\
    $\mathbf{v_i}$ & vector  & velocity of node i\\
    $V_p$ & scalar  & volume of particle p\\
    $\mathbf{P}$ & matrix  & First Piola-Kirchoff Stress\\
    $\mu, \lambda$ & scalar  & lame parameters\\
    $\mathrm{E}$ & scalar  & Young's modulus\\
    $\mathrm{\nu}$ & scalar  & Poisson's ratio\\
    \hline
    \end{tabular}
    \label{tab:mpm_notations}
\end{table}

\begin{enumerate}
    \item P2G (Particle-to-Grid): At this stage, the momentum and mass of each particle are scattered to the neighbouring grid nodes. 
    \begin{align}
        (m\mathbf{v})_i&=\sum_{p}w_{ip}[m_{p}\mathbf{v_p}+m_{p}\mathbf{C_p}(\mathbf{x_i}-\mathbf{x_p})+\mathbf{f_{ip}}\triangle t],\\
        m_i&=\sum_{p}w_{ip}m_p.
    \end{align}
    The stress $\mathbf{f_{ip}}$ between particle and grid node is computed as follows:
    \begin{align}
        \mathbf{f_{ip}}&=-\frac{4}{\triangle x^2}V_{p}^{0}\mathbf{P(F_{p}^{'})}\mathbf{F_{p}^{T}}(\mathbf{x_i}-\mathbf{x_p}),\label{equ::fip}\\
        \mathbf{F_{p}^{'}}&=(1+\triangle t\mathbf{C_p})\mathbf{F_{p}},
    \end{align}
    where $\mathbf{F_{p}^{'}}$ is the updated deformation gradient. The first piola-kirchoff stress of Neo-Hookean model is computed as follows:
    \begin{equation}
        \mathbf{P}=\mu(\mathbf{F}-\mathbf{F^{T}})+\lambda log(J)\mathbf{F^{-T}},
    \end{equation}
    where $\mu$ and $\lambda$ are lame parameters, they are computed by using Young's modulus $\mathrm{E}$ and Poisson's ratio $\mathrm{\nu}$:
    \begin{align}
        \mu=\frac{\mathrm{E}}{2(1+\mathrm{\nu})}
        , \,\lambda=\frac{\mathrm{E}\mathrm{\nu}}{(1+\mathrm{\nu})(1-2\mathrm{\nu})}.
    \end{align}
    Young's modulus is a critical parameter used to characterize the stiffness of the soft tissue.
    We will test the simulation results of manipulation of deformable tissue with different Young's modulus in \ref{RESULTS}.
    \item Grid Operation: After P2G, each grid node has gathered the momentum and mass from the surrounding particles, so the velocity of grid node can be calculated:
    \begin{equation}
        \mathbf{v_i}=\frac{(m\mathbf{v})_i}{m_i}.
    \end{equation}
    In this step, we handle the constraints and collisions. 
    This may change grid nodes' velocities to avoid penetration and realize interaction between soft tissue and surgical robot. 
    As shown in Fig. \ref{fig::sys_design}, the values of two nodes are set to zero (white color) in Grid Operation stage of MPM algorithm.
    This operation is often used to prevent particles from going out of bounds.
    The implementation details about collision detection and handling will be illustrated at \ref{collision}.
    Performing collision detection and handling on grid nodes is a more efficient strategy in contrast to particle-level handling. The enhanced efficiency stems from two key factors. Firstly, the relatively lower number of grid nodes, compared to particles, minimizes computational overhead. Secondly, the well-organized memory layout of grid nodes data enables efficient data access, thereby further optimizing computational performance.
    \item G2P (Grid-to-Particle) and Advection: In this step, we scatter the velocity of grid nodes to surrounding particles,
    \begin{equation}
        \mathbf{v_p}=\sum_{i}w_{ip}\mathbf{v_i},
    \end{equation}
    update the velocity gradient
    \begin{equation}
    \mathbf{C_p^{'}}=\frac{4}{\triangle x^2}\sum_{i}w_{ip}\mathbf{v_i}(\mathbf{x_i}-\mathbf{x_p})^T,
    \end{equation}
    and move the particles according to their velocities,
    \begin{equation}
        \mathbf{x_p^{'}}=\mathbf{x_p}+\triangle t\mathbf{v_p}.
    \end{equation}
    After advection, one simulation step is finished and the algorithm will start next simulation step from P2G again until we stop the simulation loop. 
\end{enumerate}

The combination of Lagrangian particles and Eulerian grids in MPM allows for the advantages of both approaches. 
Lagrangian particles accurately represent the motion and deformation of soft tissue, while Eulerian grids provide a convenient framework for solving the governing equations on a fixed grid. 
The coupling between Lagrangian and Eulerian representations enables the simulation of complex tissue behavior in a computationally efficient manner.

\subsection{Collision Detection and Handling}
\label{collision}
The contact and interaction between surgical tools and soft tissue frequently occur during surgeries, highlighting the significance of collision detection and handling in medical simulation.
The rigid body simulations, including the motion of the surgical robot and surgical tools, are performed within the physics engine of PyBullet\cite{PyBullet}.
The soft tissue simulation is a new developed module by ourselves.
The collision detection and handling plays a crucial role in connecting the rigid simulation and soft simulation so that these two modules are able to influence each other.
We implemented collision detection and handling to couple the soft body and rigid body using a method inspired by \cite{mpm_course}. 

In MPM algorithm, collisions are processed at the Grid Operation stage which reduces computational overhead and ensures the velocity continuity at the same time.
The key to implement collision detection lies in efficiently computing the distance between grid nodes and surgical tools.
We utilize two different methods to compute the distance according to the geometry complexity of the collision objects.
For simple objects such as the gripper of surgical robot, we estimate the appearance using two bounding boxes.
The computation of the distance between a point in space and a box surface is straightforward and can be easily accomplished.
For objects with complex geometry details, we compute the signed distance field (SDF) of them and query the distance from grid nodes to the surface of object to judge whether a grid node is collided with a rigid body (distance$<$0). 
To prevent penetration, we set a threshold $\theta$ which is close to zero, and collision will be detected if distance$<\theta$.

Recomputing SDF at each simulation step is impractical for interactive application because of the high computation cost. 
In our implementation, we only calculate SDF once and this calculation can be pre-computed so we don't need spend any time on SDF computation during the simulation, greatly improving the performance of program.
As a result, our platform is capable of simulating soft bodies at interactive speeds.
The main idea of this method is switching grid nodes from the world frame to collision objects' reference frame. 
Orientation $\mathbf{R}$ and translation $\mathbf{T}$ of the surgical tools can be gotten from SurRoL\cite{xu2021surrol}. 
The position of grid node in the reference frame of surgical tool can be computed:
\begin{equation}
    \mathbf{x_{ref}}=\mathbf{R^{-1}}(\mathbf{x}-\mathbf{T}),
\end{equation}
where $\mathbf{x}$ is the position of grid node in the world reference frame. 
After getting the position at the reference frame of rigid tool, this 3D vector can be used to index SDF (which has been pre-computed in the local reference frame of surgical tool) to query the distance between grid nodes and tools:
\begin{align}
    distance&=SDF[\frac{\mathbf{x_{ref}}[0]}{dx},\frac{\mathbf{x_{ref}[1]}}{dx},\frac{\mathbf{x_{ref}}[2]}{dx}],\\
    dx&=\frac{1}{SDF\_Resolution},\label{equ::dx}
\end{align}
where SDF\_Resolution is the resolution of SDF, and we set it to 256 in our implementation.

\begin{figure}[t]
	\centering
    \vspace*{5pt}
	\includegraphics[width=0.8\columnwidth]{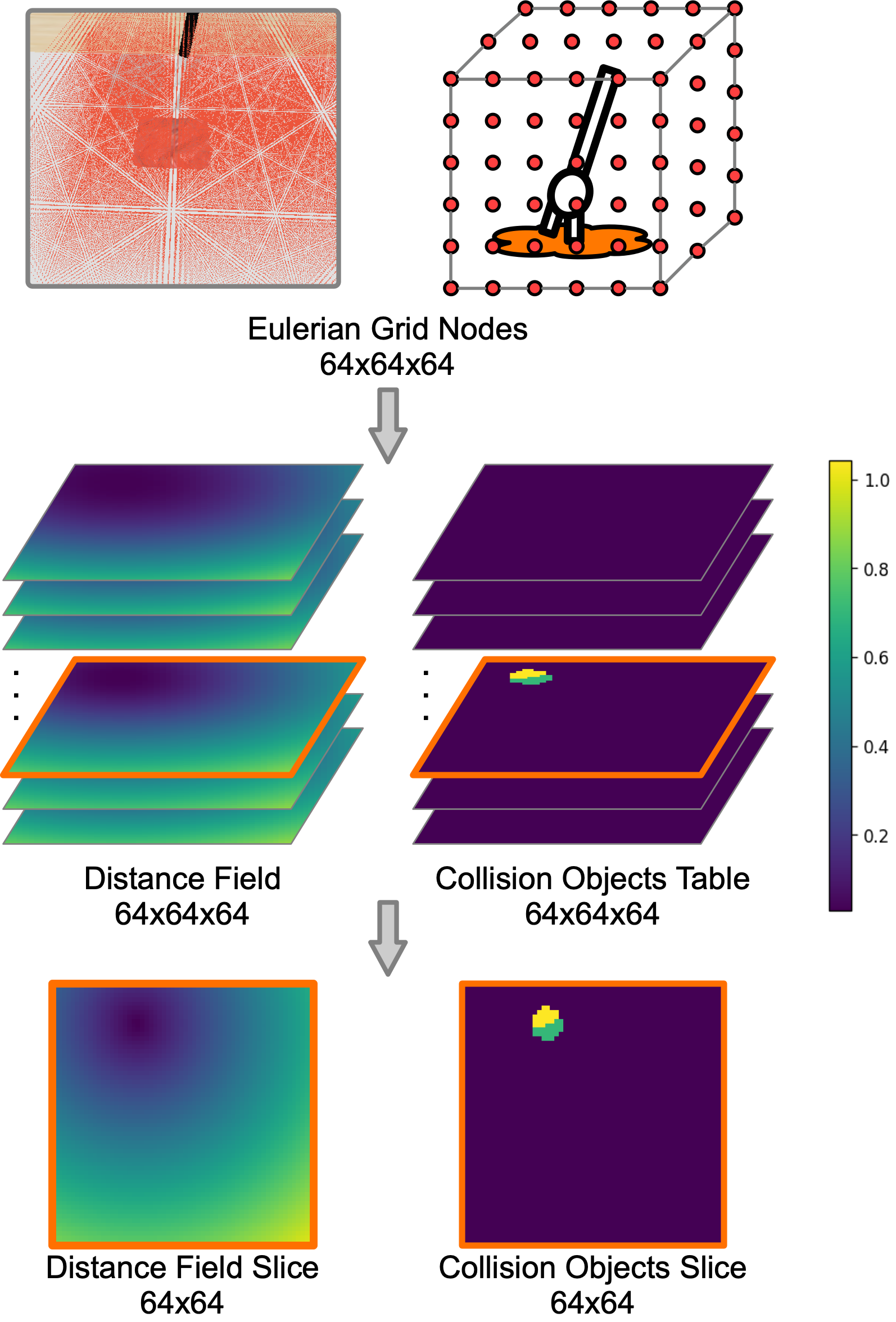}
 \caption{\textbf{Illustration of Our Collision Detection Method.} 
 The top images illustrate the relationship between Eulerian Grids nodes and simulation scene.
 At each simulation step, the distances from grid nodes to the nearest collision object and the associated object id are recorded into Distance Field and Collision Objects Table respectively.
 One distance field slice and the corresponding collision objects table slice are displayed at the lower part. 
 }
 \label{fig::collision}
\end{figure}

\begin{figure*}[t]
   \vspace*{5pt}
   \includegraphics[width=\textwidth]{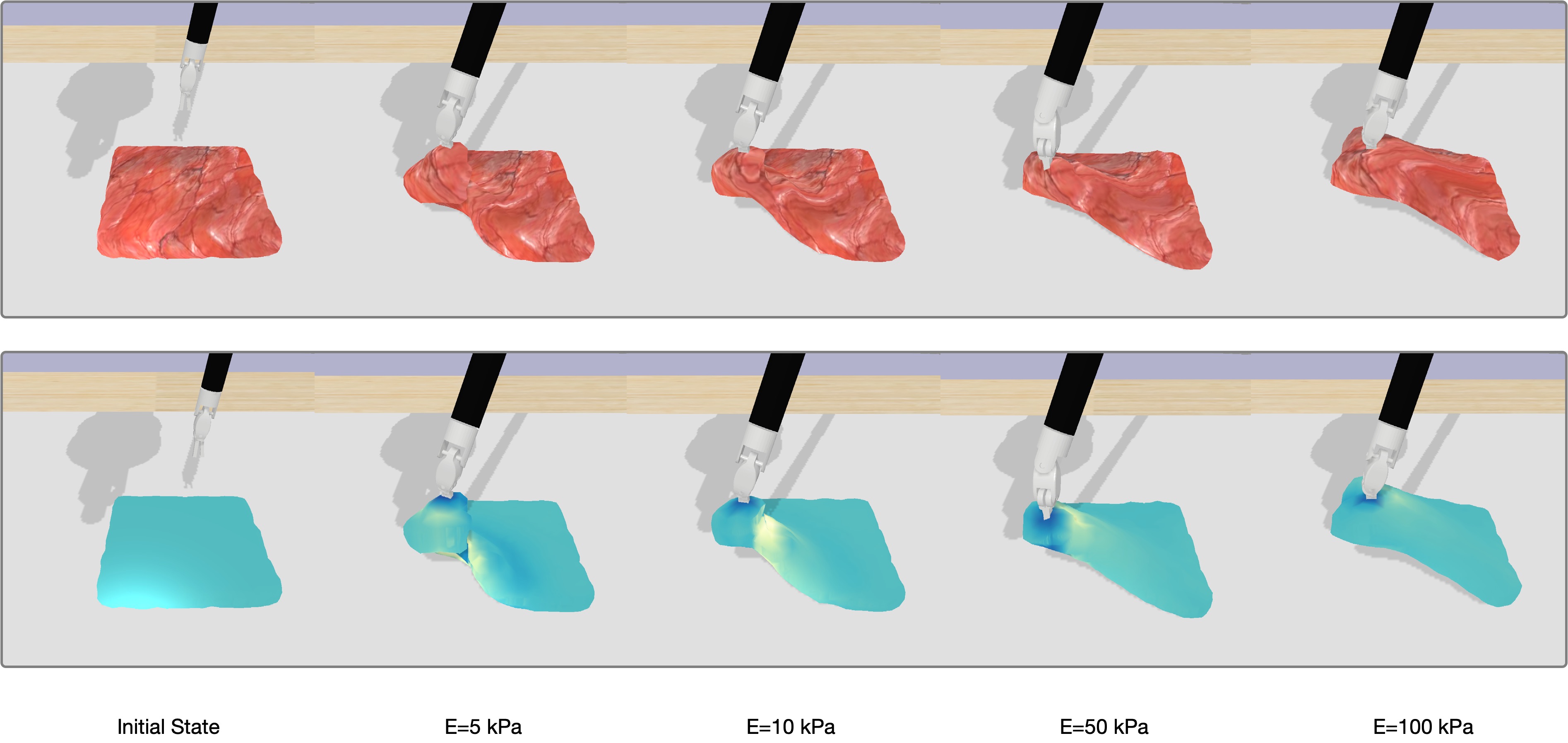}
   \caption{\textbf{Results of tissue retraction~\cite{pore2021learning,tagliabue2020soft} task under different tissue stiffness.} This image displays the simulation results of tissue retraction with four different Young's Modulus (E) values. 
   A higher Young's modulus indicates a stiffer material, while a lower value represents a more flexible or elastic material.
   All results are simulated using 24K particles and the material density was set to 1000 kg/$m^3$. 
   In the second row of the image, the color represents the degree of deformation. 
   Dark colors indicate the corresponding part of material is squeezed while light colors mean the area is being stretched.}
   \vspace{-1em}
   \label{fig::seq2}
\end{figure*}

If collision is detected at one grid node, the velocity of this node will be modified by using Coulomb friction model.
At first, the normal direction of the contact point on surgical tool can be estimated from SDF by using finite difference:
\begin{equation}
    \begin{split}
        \mathbf{n}=(&\frac{SDF[i+1,j,k]-SDF[i-1,j,k]}{2dx},\\
    &\frac{SDF[i,j+1,k]-SDF[i,j-1,k]}{2dx},\\
    &\frac{SDF[i,j,k+1]-SDF[i,j,k-1]}{2dx}),
    \end{split}
\end{equation}
where dx is same as Equation \ref{equ::dx}. Then, the grid velocity is transformed into the reference frame of the collision object,
\begin{equation}
    \mathbf{v_{rel}}=\mathbf{v}-\mathbf{v_{co}},
\end{equation}
where $\mathbf{v_{co}}$ is the velocity of collision object (surgical tool).
If grid node is approaching the collision object,
\begin{equation}
    v_n=\mathbf{v_{rel}}\cdot\mathbf{n},
\end{equation}
which means $v_n <0$, the new velocity of grid node will be computed as follows:
\begin{align}
    \mathbf{v_t}&=\mathbf{v_{rel}}-v_n\mathbf{n} \label{projection},\\
    \mathbf{v_{rel}}^{'}&=\mathbf{v_t}+\mu v_n\frac{\mathrm{v_t}}{||\mathrm{v_t}||}\label{friction},\\
    \mathbf{v^{'}}&=\mathbf{v_{rel}^{'}}+\mathbf{v_{co}}\label{pull_back_frame}.
\end{align}
Equation \ref{projection} set the relative normal velocity to zero which means the grid node and collision object share the same velocity at normal direction, which is important to handle the collision between soft tissue and surgical tools.
Equation \ref{friction} modifies the tangential velocity according to Coulomb model to simulate the friction phenomenon. 
Finally, Equation \ref{pull_back_frame} switches the processed velocity from relative reference frame to the world frame.
We use collision objects table (as shown in Fig. \ref{fig::collision}) to distinguish which object is collided with a specific grid node.

In medical scenarios, a biological tissue or organ is often manipulated by multiple surgical tools such as forceps and retractor at the same time.
To enable our simulator to detect collisions with multiple rigid bodies, we maintain the distance between grid nodes and collision objects by using the methodology inspired by the depth buffer (or z-buffer) used in computer graphics.  
The distance from grid nodes to each collision object is computed and finally merged into one array by comparing the values at the same location.
An additional array named ``Collision Objects Table" (as shown in Fig.~\ref{fig::collision}) is added to record the object id to distinguish which object is collided with a specific grid node. 
For example, in Fig.~\ref{fig::collision}, the yellow and green colors in the Collision Objects Slice represent two different jaws of the surgical robot gripper that collided with the grid nodes.
At the collision detection stage, we go through each value of the merged array and check whether the value is less than the threshold we set to detect collision, if collision is detected, the simulator will take the object id at the same position in collision objects table to get the velocity of collision object and perform collision handling.

\section{Results}\label{RESULTS}
In this section, we present the numerical and visual results of soft body simulation in our embodied environment.
We also analyze the simulation performance and factors influencing performance in \ref{performance}.
Please refer to the supplementary video for dynamic simulation results and performance comparison.

\subsection{Soft Body Simulation Results}
To test the visual effects of soft tissue simulation, we conducted different interactions with a soft tissue using surgical robot. 
Fig. \ref{fig::teaser} presents the motion sequences of 4 commonly encountered operations in surgeries: pushing\cite{marban2019recurrent}, pulling\cite{scheikl2022sim}, tearing\cite{wolper2020anisompm} and
    tensioning\cite{ou2023sim}. 
Fig. \ref{fig::seq2} displays the simulation results of tissue retraction task using materials of varying stiffness which further validates the physical interpretability of our simulator.
As illustrated in Fig. \ref{fig::seq2}, the tissue characterized by a smaller Young's Modulus (indicating a softer material) demonstrates significant deformation, with only a small portion of the material detaching from the table.
On the contrary, the harder material exhibits smaller deformation, indicating the stronger ability to maintain the original shape.
The variation in simulation results observed with different physical parameters are conformed to the physical laws, which highlights the physically-based nature of our simulator.
As displayed in Fig. \ref{fig::teaser} and Fig. \ref{fig::seq2} 
, our simulator possesses the capability to simulate the manipulation of soft bodies, and can be controlled with physical parameters in a physically interpretable manner.
These characteristics of our simulator are helpful to simulate a wide range of surgical tasks that involve the soft~tissue~manipulation.

\subsection{Simulation Performance}\label{performance}
\begin{figure}[t]
    \centering
    \vspace*{5pt}
    \includegraphics[width=\columnwidth]{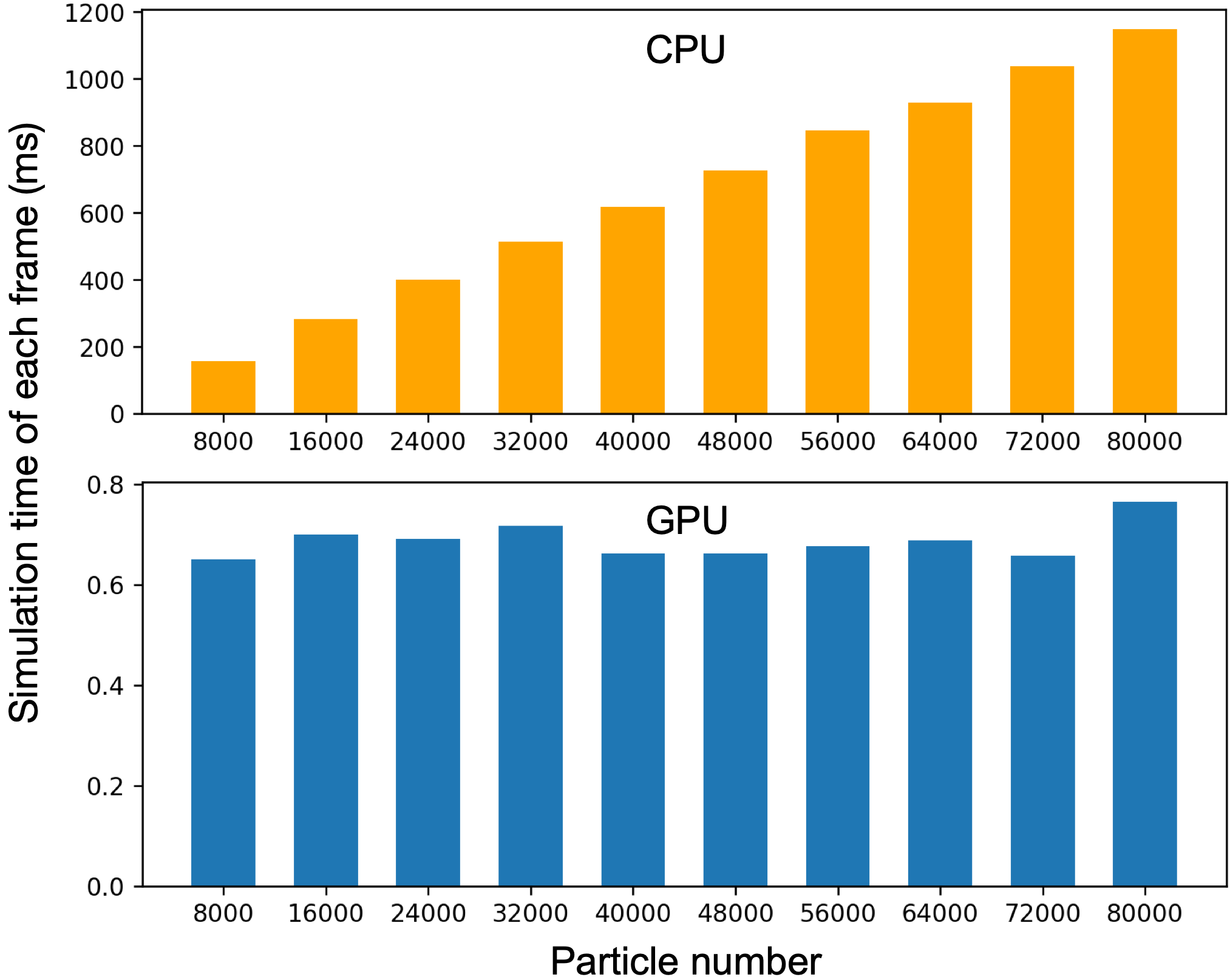}
    \caption{\textbf{Performance Statistics.}  This figure illustrates the simulation time of each frame with various simulation scales and hardware. 
    All simulation results are generated using $64\times64\times64$ grid resolution and each frame consists of 25 sub-steps. 
    The timestep of each sub-step was set to 0.0005s.}
    \label{fig::performance}
\end{figure}

Thanks to Taichi programming language\cite{taichi}, our soft body simulator has been greatly optimized and accelerated with the help of GPU. 
The simulation efficiencies at different simulation scales are illustrated in Fig. \ref{fig::performance}. 
All results are obtained from a computer equipped with an Intel i7-1165G7 CPU and an NVIDIA GeForce RTX 2060 GPU. 
Each simulation step comprises multiple sub-steps which serve as the basic units of simulation. 
In our implementation, a step consists of 25 sub-steps and each sub-step corresponds to 0.0005s.
Choosing an appropriate timestep is crucial, as it balances precision and computational cost.
Smaller timesteps offer higher precision but require more computation, while larger timesteps provide higher efficiency at the cost of reduced precision and potential instability in simulation results. 
The selection of the timestep value (0.0005s) was based on experimental results. 
As shown in Fig. \ref{fig::performance}, for simulation performed on CPU, the soft body simulation time increases linearly with the number of particles. 
In contrast, with the help of the strong parallel processing capability of GPU, the soft body simulation time is not sensitive to the number of particles and the simulation time can be dramatically reduced ($\sim \mathbf{1500}\times$ faster).

Our soft body simulator is able to execute stably at around \textbf{47 FPS} on our machine.
The computational time of each component within a single simulation step is illustrated in Table \ref{fig::frame_proportion}.
The results in the Table \ref{fig::frame_proportion} are the average of 1K simulation frames using 24K particles.
The time spent on each frame can be mainly divided into six parts. 
The collision detection, soft body simulation and rigid body simulation are directly related to simulation.
The Marching cubes, data transfer and PyBullet Visualization pertain to visualization.
Notably, the simulation part accounts for a relatively small proportion, approximately 24\%, while the visualization is the bottleneck to further improve the FPS.
The long time spent on PyBullet Visualization mainly arises from the frequent construction and deconstruction of objects required for visualizing deformable objects in PyBullet.
This issue is caused by the unfixed vertices number of the extracted surface from mass volume by using Marching Cubes\cite{marching_cubes}.
The proportion of data transfer between CPU and GPU is the largest within a simulation step and increases with the particle number.
In this paper, we focus on introducing physically based soft body simulation method into SurRoL simulation platform and improving the simulation speed by utilizing hardware acceleration.
How to efficiently visualize the deformable objects still needs further exploration.

\begin{table}[t]
    \centering
    \vspace*{5pt}
    \caption{\textbf{Computational Time of Each Component for One Step.}}
    \begin{tabular}{|l|c|c|}
    \hline
     &\textbf{Time Usage}  &\textbf{Proportion}\\
    \hline
    PyBullet Visualization&3.36ms&16.00\%\\
    Data Transfer &9.83ms&46.81\%\\
    Marching Cubes & 2.53ms&12.05\%\\ \hline
    Others&0.29ms&1.38\%\\ \hline
    Collision Detection& 3.77ms&17.95\%\\
    Rigid Simulation& 0.59ms&2.81\%\\
    Soft Simulation& \textbf{0.63ms} &3.00\%\\ \hline
    Sum&21.0ms&100.00\%\\
    \hline
    \end{tabular}
    \vspace{-1em}
    \label{fig::frame_proportion}
\end{table}
\section{Conclusion and Discussion}
In this paper, we extended an existing surgical robot simulation platform SurRoL by introducing physically based soft body simulation. 
The extended simulator provides an embodied environment which can be used to efficiently generate physically based training data of surgical tasks involving soft tissue manipulation.
Our newly extended platform paves the way for the development of automation of surgical tasks related to soft tissue manipulation.
There are some questions which are worthy of further exploration.
For example, Marching Cubes algorithm\cite{marching_cubes} can't fix the vertices number of the extracted surface from the volume data across different stages of the simulation, which poses challenges in texture mapping and also improves the visualization cost.
Furthermore, the physical parameters we used in simulation are manually curated which may influence the visual realism.
Differentiable physics simulation and learning technique may help find more appropriate parameters\cite{li2023pac,su2023generalized}.
We are committed to addressing these challenges and improving our embodied environment.
Hope our newly extended SurRoL platform can bring convenience to researchers who are interested in surgical robot and medical~simulation.
\label{CONCLUSIONS}










\bibliographystyle{IEEEtran}
\bibliography{IEEEabrv,ref}

\begin{thebibliography}{10}
\providecommand{\url}[1]{#1}
\csname url@rmstyle\endcsname
\providecommand{\newblock}{\relax}
\providecommand{\bibinfo}[2]{#2}
\providecommand\BIBentrySTDinterwordspacing{\spaceskip=0pt\relax}
\providecommand\BIBentryALTinterwordstretchfactor{4}
\providecommand\BIBentryALTinterwordspacing{\spaceskip=\fontdimen2\font plus
\BIBentryALTinterwordstretchfactor\fontdimen3\font minus \fontdimen4\font\relax}
\providecommand\BIBforeignlanguage[2]{{%
\expandafter\ifx\csname l@#1\endcsname\relax
\typeout{** WARNING: IEEEtran.bst: No hyphenation pattern has been}%
\typeout{** loaded for the language `#1'. Using the pattern for}%
\typeout{** the default language instead.}%
\else
\language=\csname l@#1\endcsname
\fi
#2}}

\bibitem{scheikl2022sim}
P.~M. Scheikl, E.~Tagliabue, B.~Gyenes, M.~Wagner, D.~Dall'Alba, P.~Fiorini, and F.~Mathis-Ullrich, ``Sim-to-real transfer for visual reinforcement learning of deformable object manipulation for robot-assisted surgery,'' \emph{IEEE Robotics and Automation Letters}, vol.~8, no.~2, pp. 560--567, 2022.

\bibitem{marban2019recurrent}
A.~Marban, V.~Srinivasan, W.~Samek, J.~Fern{\'a}ndez, and A.~Casals, ``A recurrent convolutional neural network approach for sensorless force estimation in robotic surgery,'' \emph{Biomedical Signal Processing and Control}, vol.~50, pp. 134--150, 2019.

\bibitem{wolper2020anisompm}
J.~Wolper, Y.~Chen, M.~Li, Y.~Fang, Z.~Qu, J.~Lu, M.~Cheng, and C.~Jiang, ``Anisompm: Animating anisotropic damage mechanics: Supplemental document,'' \emph{ACM Trans. Graph}, vol.~39, no.~4, 2020.

\bibitem{ou2023sim}
Y.~Ou and M.~Tavakoli, ``Sim-to-real surgical robot learning and autonomous planning for internal tissue points manipulation using reinforcement learning,'' \emph{IEEE Robotics and Automation Letters}, vol.~8, no.~5, pp. 2502--2509, 2023.

\bibitem{xu2021surrol}
J.~Xu, B.~Li, B.~Lu, Y.-H. Liu, Q.~Dou, and P.-A. Heng, ``Surrol: An open-source reinforcement learning centered and dvrk compatible platform for surgical robot learning,'' in \emph{2021 IEEE/RSJ International Conference on Intelligent Robots and Systems (IROS)}.\hskip 1em plus 0.5em minus 0.4em\relax IEEE, 2021, pp. 1821--1828.

\bibitem{long2022integrating}
Y.~Long, J.~Cao, A.~Deguet, R.~H. Taylor, and Q.~Dou, ``Integrating artificial intelligence and augmented reality in robotic surgery: An initial dvrk study using a surgical education scenario,'' 2022.

\bibitem{long2023humanintheloop}
Y.~Long, W.~Wei, T.~Huang, Y.~Wang, and Q.~Dou, ``Human-in-the-loop embodied intelligence with interactive simulation environment for surgical robot learning,'' 2023.

\bibitem{huang2023demonstrationguided}
T.~Huang, K.~Chen, B.~Li, Y.-H. Liu, and Q.~Dou, ``Demonstration-guided reinforcement learning with efficient exploration for task automation of surgical robot,'' 2023.

\bibitem{huang2023valueinformed}
T.~Huang, K.~Chen, W.~Wei, J.~Li, Y.~Long, and Q.~Dou, ``Value-informed skill chaining for policy learning of long-horizon tasks with surgical robot,'' 2023.

\bibitem{gurtin1982introduction}
M.~E. Gurtin, \emph{An introduction to continuum mechanics}.\hskip 1em plus 0.5em minus 0.4em\relax Academic press, 1982.

\bibitem{dVRL}
F.~Richter, R.~K. Orosco, and M.~C. Yip, ``Open-sourced reinforcement learning environments for surgical robotics,'' \emph{arXiv preprint arXiv:1903.02090}, 2019.

\bibitem{dvrk}
P.~Kazanzides, Z.~Chen, A.~Deguet, G.~S. Fischer, R.~H. Taylor, and S.~P. DiMaio, ``An open-source research kit for the da vinci{\textregistered} surgical system,'' in \emph{2014 IEEE international conference on robotics and automation (ICRA)}.\hskip 1em plus 0.5em minus 0.4em\relax IEEE, 2014, pp. 6434--6439.

\bibitem{munawar2019asynchronous}
A.~Munawar and G.~S. Fischer, ``An asynchronous multi-body simulation framework for real-time dynamics, haptics and learning with application to surgical robots,'' in \emph{2019 IEEE/RSJ International Conference on Intelligent Robots and Systems (IROS)}.\hskip 1em plus 0.5em minus 0.4em\relax IEEE, 2019, pp. 6268--6275.

\bibitem{ambf}
A.~Munawar, N.~Srishankar, and G.~S. Fischer, ``An open-source framework for rapid development of interactive soft-body simulations for real-time training,'' in \emph{2020 IEEE International Conference on Robotics and Automation (ICRA)}, 2020, pp. 6544--6550.

\bibitem{suturing_ambf}
A.~Munawar, J.~Y. Wu, G.~S. Fischer, R.~H. Taylor, and P.~Kazanzides, ``Open simulation environment for learning and practice of robot-assisted surgical suturing,'' \emph{IEEE Robotics and Automation Letters}, vol.~7, no.~2, pp. 3843--3850, 2022.

\bibitem{PyBullet}
E.~Coumans and Y.~Bai, ``Pybullet, a python module for physics simulation for games, robotics and machine learning,'' \url{http://pybullet.org}, 2016--2021.

\bibitem{MULLER_pbd}
\BIBentryALTinterwordspacing
M.~Müller, B.~Heidelberger, M.~Hennix, and J.~Ratcliff, ``Position based dynamics,'' \emph{Journal of Visual Communication and Image Representation}, vol.~18, no.~2, pp. 109--118, 2007. [Online]. Available: \url{https://www.sciencedirect.com/science/article/pii/S1047320307000065}
\BIBentrySTDinterwordspacing

\bibitem{tagliabue-sim}
E.~Tagliabue, A.~Pore, D.~Dall’Alba, E.~Magnabosco, M.~Piccinelli, and P.~Fiorini, ``Soft tissue simulation environment to learn manipulation tasks in autonomous robotic surgery,'' in \emph{2020 IEEE/RSJ International Conference on Intelligent Robots and Systems (IROS)}, 2020, pp. 3261--3266.

\bibitem{macklin2014unified}
M.~Macklin, M.~M{\"u}ller, N.~Chentanez, and T.-Y. Kim, ``Unified particle physics for real-time applications,'' \emph{ACM Transactions on Graphics (TOG)}, vol.~33, no.~4, pp. 1--12, 2014.

\bibitem{SULSKY_mpm}
\BIBentryALTinterwordspacing
D.~Sulsky, S.-J. Zhou, and H.~L. Schreyer, ``Application of a particle-in-cell method to solid mechanics,'' \emph{Computer Physics Communications}, vol.~87, no.~1, pp. 236--252, 1995, particle Simulation Methods. [Online]. Available: \url{https://www.sciencedirect.com/science/article/pii/0010465594001707}
\BIBentrySTDinterwordspacing

\bibitem{taichi}
Y.~Hu, T.-M. Li, L.~Anderson, J.~Ragan-Kelley, and F.~Durand, ``Taichi: a language for high-performance computation on spatially sparse data structures,'' \emph{ACM Transactions on Graphics (TOG)}, vol.~38, no.~6, p. 201, 2019.

\bibitem{macklin2016xpbd}
M.~Macklin, M.~M{\"u}ller, and N.~Chentanez, ``Xpbd: position-based simulation of compliant constrained dynamics,'' in \emph{Proceedings of the 9th International Conference on Motion in Games}, 2016, pp. 49--54.

\bibitem{yin2021modeling}
H.~Yin, A.~Varava, and D.~Kragic, ``Modeling, learning, perception, and control methods for deformable object manipulation,'' \emph{Science Robotics}, vol.~6, no.~54, p. eabd8803, 2021.

\bibitem{in2fem}
J.~N. Reddy, \emph{Introduction to the finite element method}.\hskip 1em plus 0.5em minus 0.4em\relax McGraw-Hill Education, 2019.

\bibitem{FLIP}
\BIBentryALTinterwordspacing
J.~Brackbill and H.~Ruppel, ``Flip: A method for adaptively zoned, particle-in-cell calculations of fluid flows in two dimensions,'' \emph{Journal of Computational Physics}, vol.~65, no.~2, pp. 314--343, 1986. [Online]. Available: \url{https://www.sciencedirect.com/science/article/pii/0021999186902111}
\BIBentrySTDinterwordspacing

\bibitem{stomakhin-snow-mpm}
A.~Stomakhin, C.~Schroeder, L.~Chai, J.~Teran, and A.~Selle, ``A material point method for snow simulation,'' \emph{ACM Transactions on Graphics (TOG)}, vol.~32, no.~4, pp. 1--10, 2013.

\bibitem{mls-mpm}
\BIBentryALTinterwordspacing
Y.~Hu, Y.~Fang, Z.~Ge, Z.~Qu, Y.~Zhu, A.~Pradhana, and C.~Jiang, ``A moving least squares material point method with displacement discontinuity and two-way rigid body coupling,'' \emph{ACM Trans. Graph.}, vol.~37, no.~4, jul 2018. [Online]. Available: \url{https://doi.org/10.1145/3197517.3201293}
\BIBentrySTDinterwordspacing

\bibitem{jiang2015affine}
C.~Jiang, C.~Schroeder, A.~Selle, J.~Teran, and A.~Stomakhin, ``The affine particle-in-cell method,'' \emph{ACM Transactions on Graphics (TOG)}, vol.~34, no.~4, pp. 1--10, 2015.

\bibitem{bar2006simbionix}
S.~Bar-Meir, ``Simbionix simulator,'' \emph{Gastrointestinal Endoscopy Clinics}, vol.~16, no.~3, pp. 471--478, 2006.

\bibitem{vrep}
G.~A. Fontanelli, M.~Selvaggio, M.~Ferro, F.~Ficuciello, M.~Vendittelli, and B.~Siciliano, ``A v-rep simulator for the da vinci research kit robotic platform,'' in \emph{2018 7th IEEE International Conference on Biomedical Robotics and Biomechatronics (Biorob)}, 2018, pp. 1056--1061.

\bibitem{fan2022unity}
K.~Fan, A.~Marzullo, N.~Pasini, A.~Rota, M.~Pecorella, G.~Ferrigno, and E.~De~Momi, ``A unity-based da vinci robot simulator for surgical training,'' in \emph{2022 9th IEEE RAS/EMBS International Conference for Biomedical Robotics and Biomechatronics (BioRob)}.\hskip 1em plus 0.5em minus 0.4em\relax IEEE, 2022, pp. 1--6.

\bibitem{surgicalgym}
S.~Schmidgall, A.~Krieger, and J.~Eshraghian, ``Surgical gym: A high-performance gpu-based platform for reinforcement learning with surgical robots,'' \emph{arXiv preprint arXiv:2310.04676}, 2023.

\bibitem{isaac}
V.~Makoviychuk, L.~Wawrzyniak, Y.~Guo, M.~Lu, K.~Storey, M.~Macklin, D.~Hoeller, N.~Rudin, A.~Allshire, A.~Handa, \emph{et~al.}, ``Isaac gym: High performance gpu-based physics simulation for robot learning,'' \emph{arXiv preprint arXiv:2108.10470}, 2021.

\bibitem{scheikl2023lapgym}
P.~M. Scheikl, B.~Gyenes, R.~Younis, C.~Haas, G.~Neumann, M.~Wagner, and F.~Mathis-Ullrich, ``Lapgym--an open source framework for reinforcement learning in robot-assisted laparoscopic surgery,'' \emph{arXiv preprint arXiv:2302.09606}, 2023.

\bibitem{sofa}
F.~Faure, C.~Duriez, H.~Delingette, J.~Allard, B.~Gilles, S.~Marchesseau, H.~Talbot, H.~Courtecuisse, G.~Bousquet, I.~Peterlik, \emph{et~al.}, ``Sofa: A multi-model framework for interactive physical simulation,'' \emph{Soft tissue biomechanical modeling for computer assisted surgery}, pp. 283--321, 2012.

\bibitem{marching_cubes}
\BIBentryALTinterwordspacing
W.~E. Lorensen and H.~E. Cline, ``Marching cubes: A high resolution 3d surface construction algorithm,'' in \emph{Proceedings of the 14th Annual Conference on Computer Graphics and Interactive Techniques}, ser. SIGGRAPH '87.\hskip 1em plus 0.5em minus 0.4em\relax New York, NY, USA: Association for Computing Machinery, 1987, p. 163–169. [Online]. Available: \url{https://doi.org/10.1145/37401.37422}
\BIBentrySTDinterwordspacing

\bibitem{mpm_course}
C.~Jiang, C.~Schroeder, J.~Teran, A.~Stomakhin, and A.~Selle, ``The material point method for simulating continuum materials,'' in \emph{Acm siggraph 2016 courses}, 2016, pp. 1--52.

\bibitem{pore2021learning}
A.~Pore, E.~Tagliabue, M.~Piccinelli, D.~Dall’Alba, A.~Casals, and P.~Fiorini, ``Learning from demonstrations for autonomous soft-tissue retraction,'' in \emph{2021 International Symposium on Medical Robotics (ISMR)}.\hskip 1em plus 0.5em minus 0.4em\relax IEEE, 2021, pp. 1--7.

\bibitem{tagliabue2020soft}
E.~Tagliabue, A.~Pore, D.~Dall’Alba, E.~Magnabosco, M.~Piccinelli, and P.~Fiorini, ``Soft tissue simulation environment to learn manipulation tasks in autonomous robotic surgery,'' in \emph{2020 IEEE/RSJ International Conference on Intelligent Robots and Systems (IROS)}.\hskip 1em plus 0.5em minus 0.4em\relax IEEE, 2020, pp. 3261--3266.

\bibitem{li2023pac}
X.~Li, Y.-L. Qiao, P.~Y. Chen, K.~M. Jatavallabhula, M.~Lin, C.~Jiang, and C.~Gan, ``Pac-nerf: Physics augmented continuum neural radiance fields for geometry-agnostic system identification,'' \emph{arXiv preprint arXiv:2303.05512}, 2023.

\bibitem{su2023generalized}
H.~Su, X.~Li, T.~Xue, C.~Jiang, and M.~Aanjaneya, ``A generalized constitutive model for versatile mpm simulation and inverse learning with differentiable physics,'' \emph{Proceedings of the ACM on Computer Graphics and Interactive Techniques}, vol.~6, no.~3, pp. 1--20, 2023.

\end{thebibliography}

\end{document}